\documentclass{article}

\PassOptionsToPackage{square,numbers}{natbib}

\usepackage[preprint]{neurips_2024}

\usepackage[utf8]{inputenc} 
\usepackage[T1]{fontenc}    
\usepackage{hyperref}       
\usepackage{url}            
\usepackage{booktabs}       
\usepackage{amsfonts}       
\usepackage{nicefrac}       
\usepackage{microtype}      
\usepackage{xcolor}         
\usepackage{graphicx}
\usepackage{subfigure}
\usepackage{wrapfig}

\title{Interpreting the Residual Stream of ResNet18}

\author{
  Andr\'e Longon \\
  Division of Computer Science and Engineering\\
  Louisiana State University\\
  Baton Rouge, LA 70803 \\
  \texttt{alongo6@lsu.edu} \\
}

\begin{document}

\maketitle

\begin{abstract}
  A mechanistic understanding of the computations learned by deep neural networks (DNNs) is far from complete.  In the domain of visual object recognition, prior research has illuminated inner workings of InceptionV1, but DNNs with different architectures have remained largely unexplored.  This work investigates ResNet18 with a particular focus on its residual stream, an architectural mechanism which InceptionV1 lacks.  We observe that for a given block, channel features of the stream are updated along a spectrum: either the input feature skips to the output, the block feature overwrites the output, or the output is some mixture between the input and block features.  Furthermore, we show that many residual stream channels compute scale invariant representations through a mixture of the input's smaller-scale feature with the block's larger-scale feature.  This not only mounts evidence for the universality of scale equivariance, but also presents how the residual stream further implements scale invariance.  Collectively, our results begin an interpretation of the residual stream in visual object recognition, finding it to be a flexible feature manager and a medium to build scale invariant representations.
\end{abstract}

\section{Introduction}

Deep neural networks (DNNs) lead humanity into a quandary: they learn to perform increasingly sophisticated tasks but leave us ignorant of how they ultimately solve them.  As DNNs are deployed in safety-critical applications such as autonomous driving and medical diagnostics, understanding their inner workings is imperative to better predict their strengths and limitations.  Additionally, the leading models of neural response predictions in visual \cite{Schrimpf407007} and language \cite{Schrimpf2021-di} processing are DNNs, so reverse-engineering these models may uncover the neural mechanisms from which these cognitive abilities emerge.

Toward this end, the field of mechanistic interpretability has progressed in illuminating DNN circuitry with a concentration on vision \cite{olah2020zoom} and language \cite{elhage2021mathematical} models.  In the visual domain, InceptionV1 \cite{Szegedy_2015_CVPR} is the most well-studied DNN where curve detectors \cite{cammarata2020curve}, equivariances for hue, rotation, and scale \cite{olah2020naturally}, and high-low frequency detectors \cite{schubert2021high-low} have been identified.  However, a key issue of mechanistic interpretability is \textit{universality}: to what extent do these mechanisms exist in other DNNs, such as those with different architectures or trained on different datasets?

To expand the mechanistic interpretation of visual models, we zoom into the \textit{residual stream} of ResNet18 \cite{He_2016_CVPR}, offering an empirical account of its behavior.  The residual stream, which InceptionV1 lacks, is an architectural mechanism that permits features to bypass layers of processing via their summation to the output of a downstream layer.  With ResNet18 being the smallest member of the ResNet family, it is manageable to perform a network-scale study of its stream, striking a balance between a simplistic ``toy model'' while having \(\sim\)5x less parameters than ResNet152, the largest of the family.

\begin{figure}[h]
    \centering
    \includegraphics[width=0.9\linewidth]{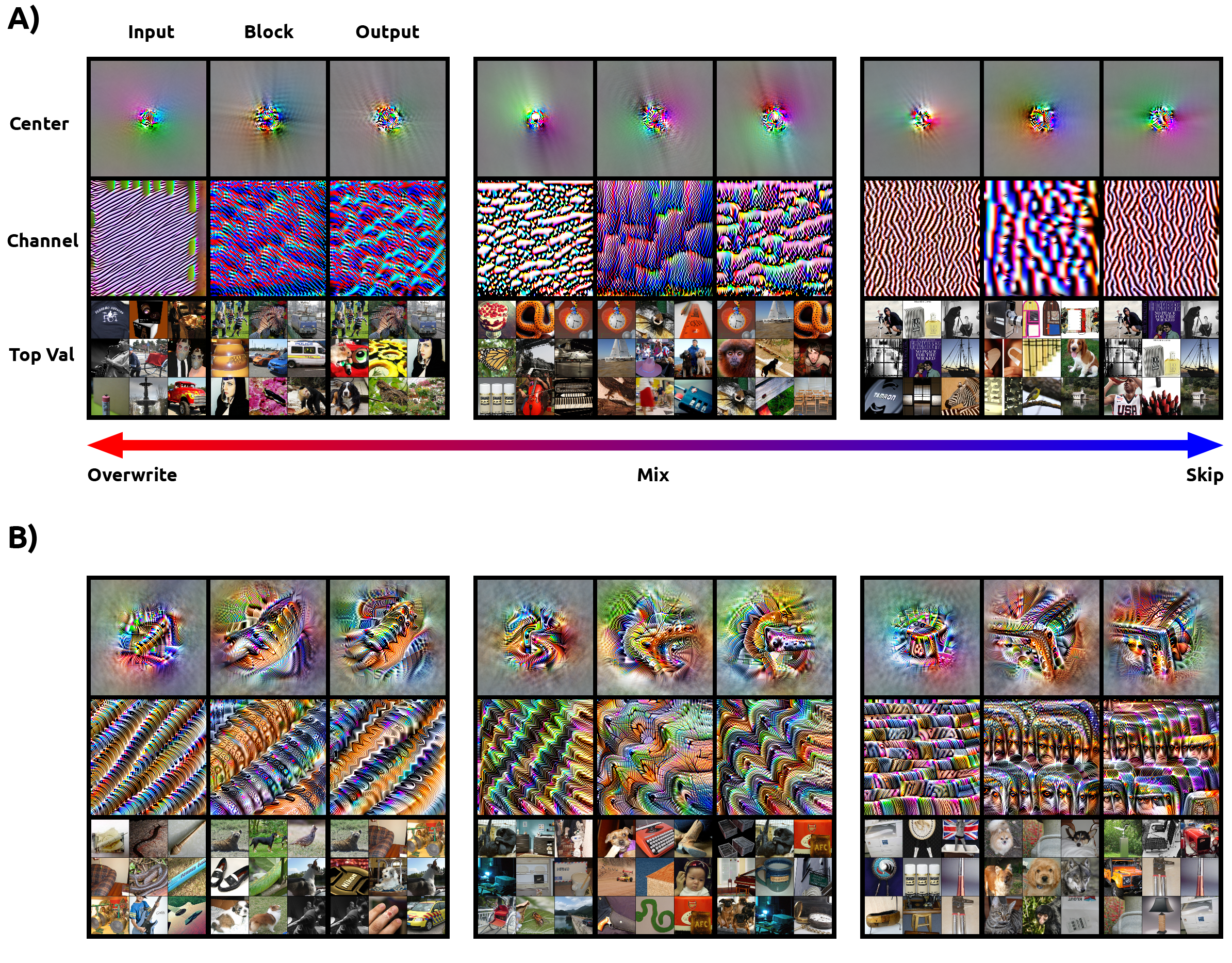}
    \caption{Each grid contains feature visualizations of the channel's center neuron, the entire channel, and the top 9 center neuron activating natural images from the ImageNet validation set \cite{5206848}.  All channel indices are zero-indexed.  \textbf{A:} Exemplary overwrite (38), mix (19), and skip (25) channels from output 1.1 (1.0 input + 1.1.bn2 block).  Note the presence or lack of overlap in top vals between Output and Input and/or Block.  \textbf{B:} Exemplary scale invariant channels (178, 215, 25) from output 3.1.}
    \label{fig:1}
\end{figure}

We present the following results from our investigation:

\begin{itemize}
    \item Stream channels at a given block update their features either by skipping the input to the output of the block, erasing the input and overwriting with the block feature if present, or mixing the input and block features (see Figure \ref{fig:1}A).  We highlight network trends of how channels are distributed across this \textit{skip-overwrite spectrum} and demonstrate some mechanisms of how the spectrum is implemented.

    \item With simple criteria, residual stream channels which exhibit \textit{scale invariance} are extracted (see Figure \ref{fig:1}B).  The input channel possesses the smaller scale version, the block's pre-sum output possesses the larger scale, and these are summed together to produce the final scale invariant output.  This pattern is especially pronounced in intermediate blocks.  These findings not only mount evidence for the universality of scale equivariance, but further show how the residual stream uses these features to construct scale invariance.
\end{itemize}

Together, our results serve as a starting point for exploring the function of the residual stream in visual processing.  We hope to inspire further work on elucidating the circuitry of these mechanisms, and even to search for similar computations in biological visual systems.  It is known that the macaque \cite{Kravitz2012-ls} and mouse \cite{Gamanut2021-az} visual systems contain so-called ``bypass connections'', where neurons project to farther downstream processing areas, skipping over intermediate areas.  This anatomical feature is highly analogous to the DNN residual stream, so it is conjectured that a function of bypass connections is, for instance, to compute scale invariance.  This possibility is important for visual neuroscience and to further establish universality in computations between artificial and biological neural networks.

\section{Methods}

We use the PyTorch \cite{NEURIPS2019_bdbca288} torchvision implementation of ResNet18 with the default ImageNet-trained \cite{5206848} weights, \verb|IMAGENET1K_V1|.  Neuron activations are obtained via the THINGSvision library \cite{Muttenthaler_2021}. 

\subsection{Residual Stream Architecture}

After an initial convolutional layer and max pool, the ResNet18 architecture is an alternating block structure of two types:  one with a downsample submodule and one without, which we refer to as a downsample block and simple block respectively.  In a simple block, its input tensor is element-wise added with the tensor obtained from its last processing step, the second batch normalization (BN).  Note there is no ReLU nonlinearity before this addition.  In a downsample block, this element-wise addition cannot occur as the block's first convolution doubles the number of channels and halves both spatial dimensions via a stride of 2.  Therefore, the input is passed through a separate convolutional layer with a stride of 2 and doubled channels, followed by a BN.  The result of this downsample submodule is then added with the second BN tensor.  In both downsample and simple blocks, the summed result undergoes a ReLU nonlinearity to produce the final output activations of the block.

\paragraph{Notation}  This work is only concerned with the activations from three layers for a given block.  We name the layers \(B\), \(I\), and \(O\) for block, input, and output respectively (the block index is not notated as we analyze them in isolation).  \(B\) is the block's second BN.  \(I\) is the layer whose output bypasses the main layers of the block and added to the output of \(B\).  For the simple block, \(I\) is the previous block's output.  In the downsample block, \(I\) is the result from the BN in the downsample submodule.  Finally, \(O\) is the layer which outputs the sum of outputs from \(I\) and \(B\) followed by a ReLU.

As an example, if the given block is simple such as 3.1, then \(B\) is the second BN in 3.1, \(I\) is the output of block 3.0, and \(O\) is the output of 3.1 itself.  For a downsample block such as 4.0, \(I\) is the BN of the downsample submodule in 4.0, \(B\) is still the second BN from 4.0, and \(O\) is still the final output of 4.0.

\subsection{Feature Visualization}

Our results rely on feature visualization, an optimization algorithm used to generate images which maximally activate a target channel (or an individual neuron) in a given layer.  Such images are henceforth referred to as feature visualizations (FZs).  We use the version of feature visualization described by \cite{olah2017feature} and implemented in the Lucent library\footnote{\url{https://github.com/greentfrapp/lucent}} for PyTorch.  We use near-identical regularizations as prescribed by \cite{olah2017feature} for improved interpretability, with altered jitter values discussed in the appendix (see Section \ref{fz_reg}).

We use a channel's center-neuron optimized FZs for all subsequent analyses.  While we also display channel-optimized FZs, they are for visualization purposes only.  We also do not study the first residual stream block (1.0) in this study.  This is due to its input resulting from a maxpool layer, making it is difficult to obtain FZs due to frequent zero-valued gradients\footnote{This problem also arises when optimizing on a post-ReLU neuron.  To circumvent this, Identity layers are added after stream summation to obtain pre-ReLU activations.}.  Therefore, we start at the 1.1 block and work our way to the end of the stream through 4.1.

\paragraph{Notation}

In a given block of the residual stream, for a given channel \(c\), we obtain the center-neuron FZs for \(I_c, B_c, \textrm{and } O_c\), which we denote as \(\hat{X}_{I_c}, \hat{X}_{B_c}, \textrm{and } \hat{X}_{O_c}\) respectively.  We denote the output of layer \(I\)'s center-neuron activation in channel \(c\) to image \(X\) as \(I_c(X)\) (equivalent notation for \(B\) and \(O\)).  Thus, \(I_c(\hat{X}_{I_c})\) is the center-neuron activation in channel \(c\) resulting from its optimized FZ.

\subsection{Mix Ratio}

In order to measure where a given block's channel falls along the Skip-Overwrite spectrum, we define the mix ratio as:
\begin{equation}
     M_c = \frac{O_c(\hat{X}_{I_c})}{O_c(\hat{X}_{B_c})}
\end{equation}
   
\(M_c\) measures the relative activations of the two components that when summed, make up \(O_c\): \(I_c\) and \(B_c\).  The mix ratio informs us of the importance of the input feature versus the block feature.  A high \(M\) thus indicates a skip-like behavior, a low \(M\) is overwrite-like, and an \(M\) close to \(1\) is a mixture of input and block features (resembling an OR gate).  We max-clamp \(M\) to \(5\) for our analyses, and as a channel's \(B_c\) can be negative, we also set those \(M_c\) values to \(5\) to indicate a skip.

\subsection{Scale Invariance Criteria} \label{scale_criteria}

We assume that if a particular \(O_c\) at a given block is to be scale invariant, then \(I_c\) will represent the smaller-scale copy, and \(B_c\) will represent the larger-scale copy.  This is a reasonable assumption as the receptive field of a neuron in \(I_c\) is always smaller than one in \(B_c\) (or equal in deep enough blocks).  We then determine if \(O_c\) exhibits scale invariance by checking if all the following conditions are true:

\begin{equation} \label{scale_eq:1}
    \frac{2}{3} < M_c < \frac{3}{2}
\end{equation}\\
\begin{equation} \label{scale_eq:2}
    \textrm{ReLU}(B_c(\hat{X}_{I_c})) < B_c(S(\hat{X}_{I_c})) 
\end{equation}\\
\begin{equation} \label{scale_eq:3}
    \textrm{ReLU}(I_c(\hat{X}_{B_c})) < I_c(S^{-1}(\hat{X}_{B_c}))
\end{equation}

where

\[
    S(X) := \textrm{Resize}(\textrm{CenterCrop(X)}) \quad \textrm{and} \quad S^{-1}(X) := \textrm{Pad}(\textrm{Resize(X)}).
\]

\(S(X)\) denotes a scale transformation of an image where it is first center-cropped to 112 pixels (half the spatial resolution), then resized back to 224 pixels using bilinear interpolation. \(S^{-1}(X)\) denotes an ``inverse''\footnote{We acknowledge that \(S^{-1}(X)\) is not a true inverse so there is an asymmetry, but we find it sufficient to provide an entry point into this study.} scale transformation where the image is resized to 112, then reflection padded to 224.

The rationale for these criteria is as follows:

\begin{itemize}
    \item Equation \ref{scale_eq:1} ensures the block output \(O_c\) is invariant (with some tolerance) to the scale of the input and block features, where both the smaller-scale \(\hat{X}_{I_c}\) and larger-scale \(\hat{X}_{B_c}\) activate \(O_c\) similarly.
    \item Equation \ref{scale_eq:2} increases the scale of the smaller-scale neuron's FZ.  It then checks if the upscaled FZ activates the larger-scale neuron more so than the unscaled FZ.  This ensures that scaling up \(\hat{X}_{I_c}\) makes it more closely resemble the larger-scale neuron's FZ.
    \item Equation \ref{scale_eq:3} similarly decreases the scale of the larger-scale neuron's FZ to see if it increases the activation of the smaller-scale neuron relative to the unscaled FZ.
\end{itemize}

We apply a ReLU in the unscaled case to guarantee we are not bringing a negative activation from the unscaled FZ up to 0 activation (or less negative) in the scaled case, indicative of destroying inhibitory selectivity without necessarily increasing the selectivity for the excitatory feature.

\paragraph{Scale Metric}  When channels that meet our criteria are identified, we next quantify the degree of their scale invariance by:

\begin{equation} \label{scale_eq:4}
    SM_c := \frac{B_c(S(\hat{X}_{I_c})) - \textrm{ReLU}(B_c(\hat{X}_{I_c}))}{B_c(\hat{X}_{B_c})} + \frac{I_c(S^{-1}(\hat{X}_{B_c})) - \textrm{ReLU}(I_c(\hat{X}_{B_c}))}{I_c(\hat{X}_{I_c})}
\end{equation}

which is the sum of activation deltas \(S(X)\) and \(S^{-1}(X)\) produce from equations \ref{scale_eq:2} and \ref{scale_eq:3}, scaled by input and block's respective FZ-evoked activations to account for different activation ranges across the channels.  We perform this scaling as this metric will be later used to rank the channels.

\section{Results}

\subsection{Skip-Overwrite Spectrum}

\begin{figure}[h]
\centering
\includegraphics[width=0.59\linewidth]{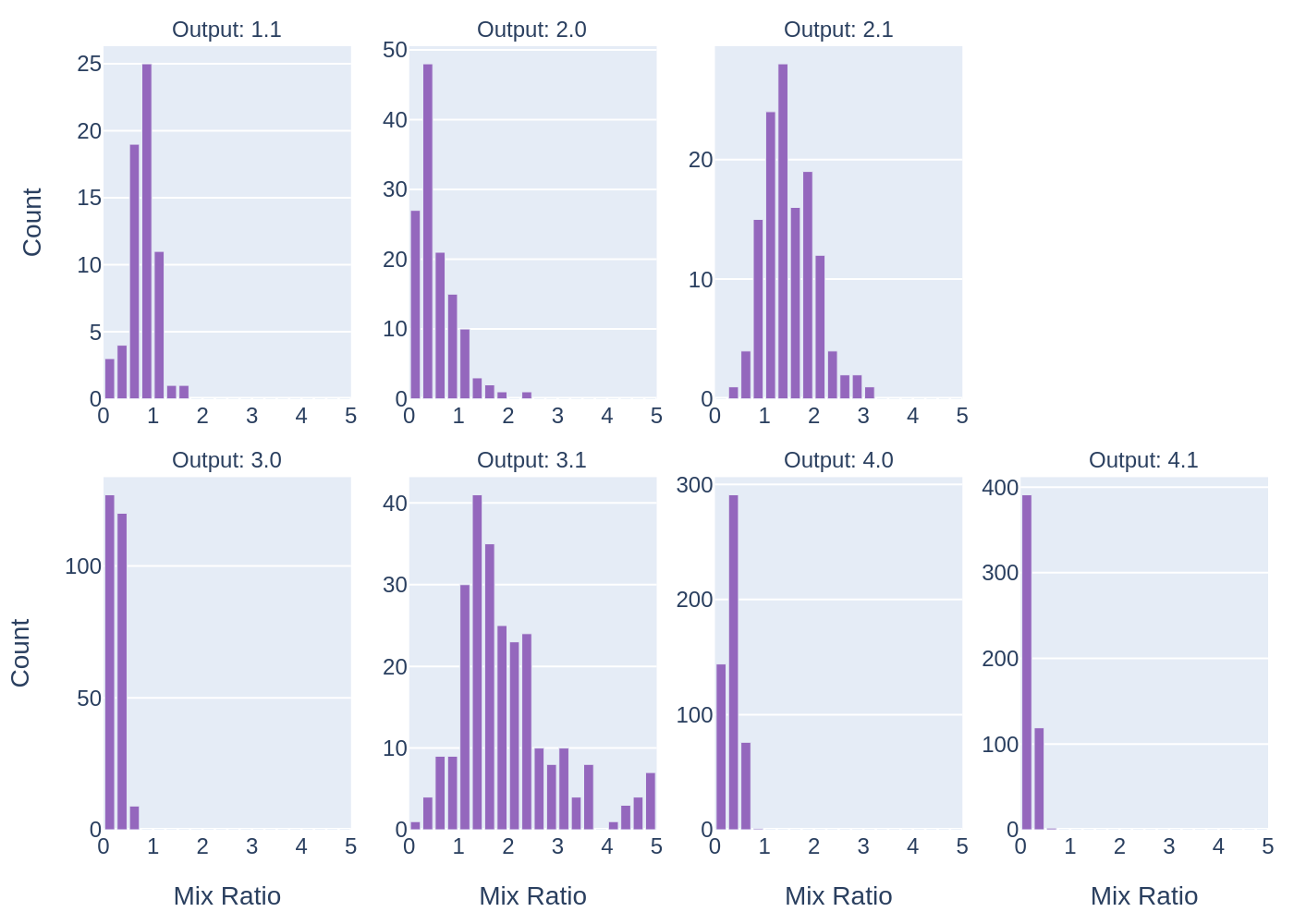}
\caption{Histograms of mix ratios for all inspected blocks of ResNet18, starting at 1.1 through 4.1.  Bins are from 0 to 5 at 0.25 intervals.  Close to 1 is a mixture of input and block features, below 1 indicates an overwrite behavior, above 1 indicates a skip behavior.}
\label{fig:2}
\end{figure}

To begin, we obtain mix ratios \(M_c\) for all channels in every block studied: 1.1 through 4.1.  As seen in Figure \ref{fig:2}, blocks exhibit a variety of distributions across the spectrum.  All downsample blocks are dominated by overwrite channels, whereas simple blocks exhibit less skewed distributions.  We note that the final block, 4.1, is an exception to this pattern: it is a simple block with all of its channels in the lowest bin (we confirm that nearly all of them are non-zero). 

\begin{figure}[h]
\centering
\includegraphics[width=0.59\linewidth]{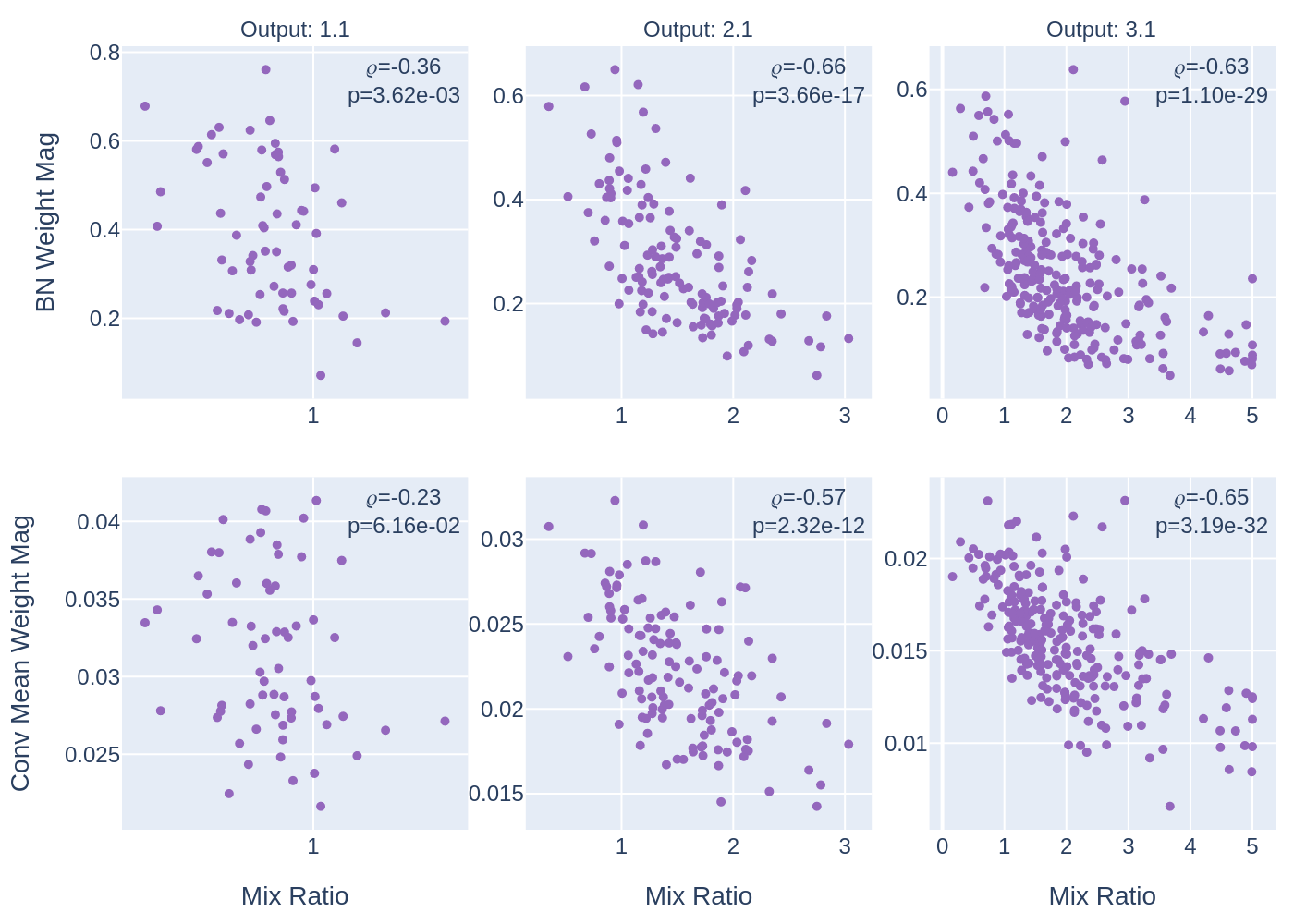}
\caption{Weight magnitudes with respect to mix ratio of second batch normalization (top row) and the second convolutional layer (bottom row, mean taken).  Spearman rank correlations and p-values displayed for each block, which show that channels with more skip-like behavior have smaller weight magnitudes.}
\label{fig:3}
\end{figure}

\paragraph{Skip Weight Analysis}  To explore the mechanisms that implement this spectrum, we first suppose that the weights of skip channels will have lower magnitude weights relative to the layer.  In Figure \ref{fig:3}, we plot each channel's second BN weight magnitude and the mean weight magnitude from their second convolutional layer.  We do not display the downsample blocks and 4.1 due to the sharp overwrite skew.  In the remaining simple blocks, we confirm that higher mix ratio channels possess significantly smaller weight magnitudes in the inspected layers.  It seems ResNet18 simply silences the block output with lower weights, allowing the input to skip to the output relatively undisturbed.

\begin{figure}[h]
\centering
\includegraphics[width=0.7\linewidth]{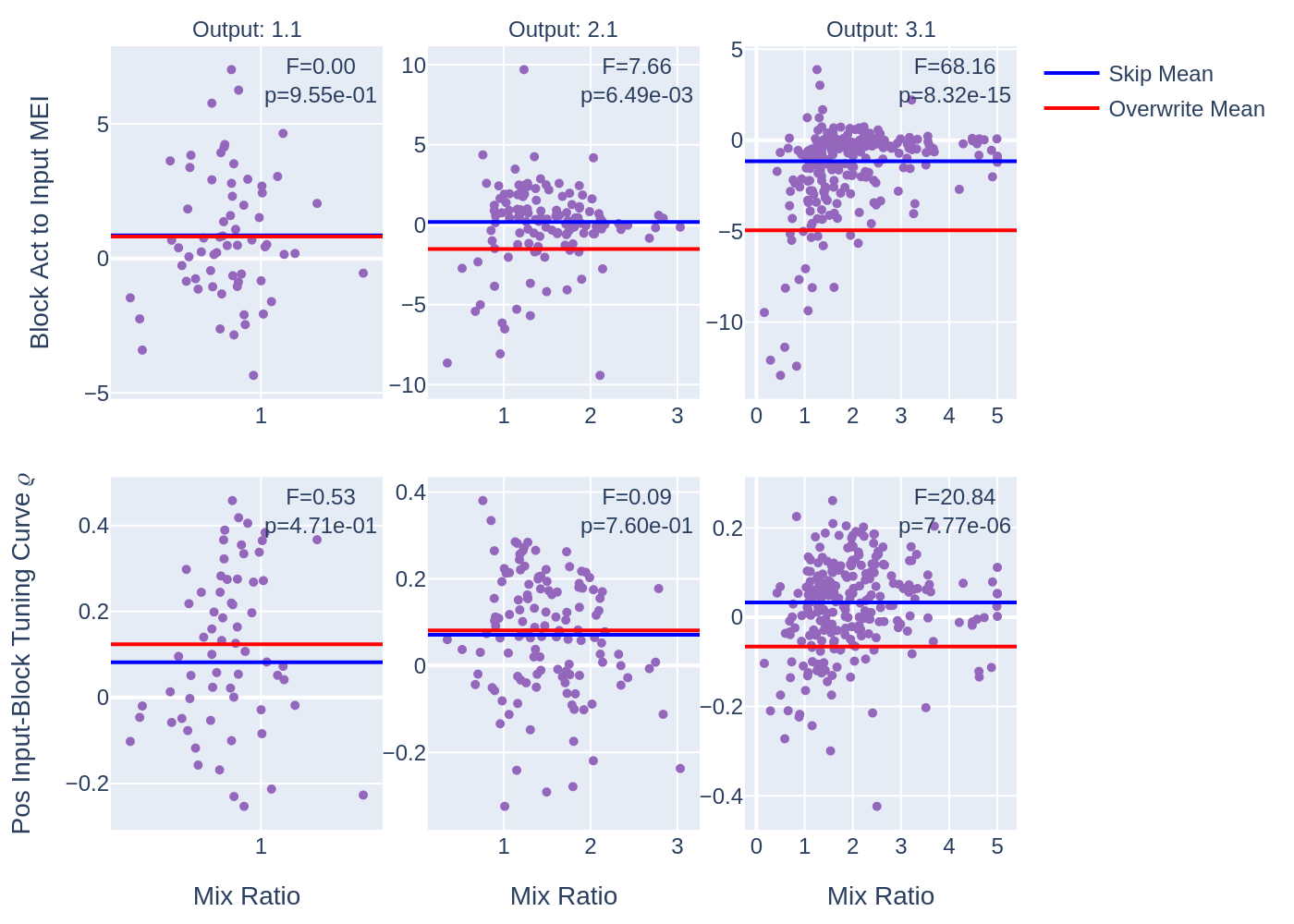}
\caption{\textbf{Top Row:}  Activations of second batch norm (pre-sum block output) with respect to mix ratio, with activation means for skip (mix > 1, blue) and overwrite (mix < 1, red) channels.  \textbf{Bottom Row:}  Spearman rank correlations of Block activations with respect to Input activations for ImageNet validation images which positively activate Input, plotted against mix ratio.  ANOVA Fs and p-values between skip and overwrite activations displayed for each block.}
\label{fig:4}
\end{figure}

\paragraph{Overwrite Inhibition}  We observe that for many channels, \(B_c(\hat{X}_{I_c}) < 0\), which appears instrumental to facilitate overwrite.  That is, when presented with the input's preferred feature, the block's second BN inhibits the output of the channel.  In Figure \ref{fig:4} (Top Row), we see an abundance of such negative activations amongst overwrite channels (\(M_c < 1\)), significantly lower than the activations of skip channels (\(M_c > 1\)) in blocks 2.1 and 3.1.  However, we find that many skip channels also possess such negative activations, so it is difficult to make conclusions from this result alone.

Next, we obtained the activations from the input and block to all ImageNet validation images where the input responded with a positive activation.  We then took a Spearman rank correlation of the block activations with respect to the input activations for each channel and report the results in \ref{fig:4} (Bottom Row), plotted against the mix ratio.  If input features inhibit the block for overwrite channels, we expect them to have significantly more negative correlations (as input activation increases, the block tends to decrease) compared to skip channels.  This is the case for block 3.1, but not 1.1 nor 2.1.

From these conflicting results, it is difficult to conclude that inhibition is a mechanism deployed to erase and/or overwrite features of the stream.  It may be that ResNet18 relies more on higher weight magnitudes to accomplish overwrite as seen in Figure \ref{fig:3}.  That said, the negative activations and correlations in block 3.1 suggest an intuitively appealing strategy by which DNNs can erase channel features and overwrite them if the block feature is present.
\begin{wrapfigure}[14]{r}{0.37\textwidth}
    \includegraphics[width=0.37\textwidth]{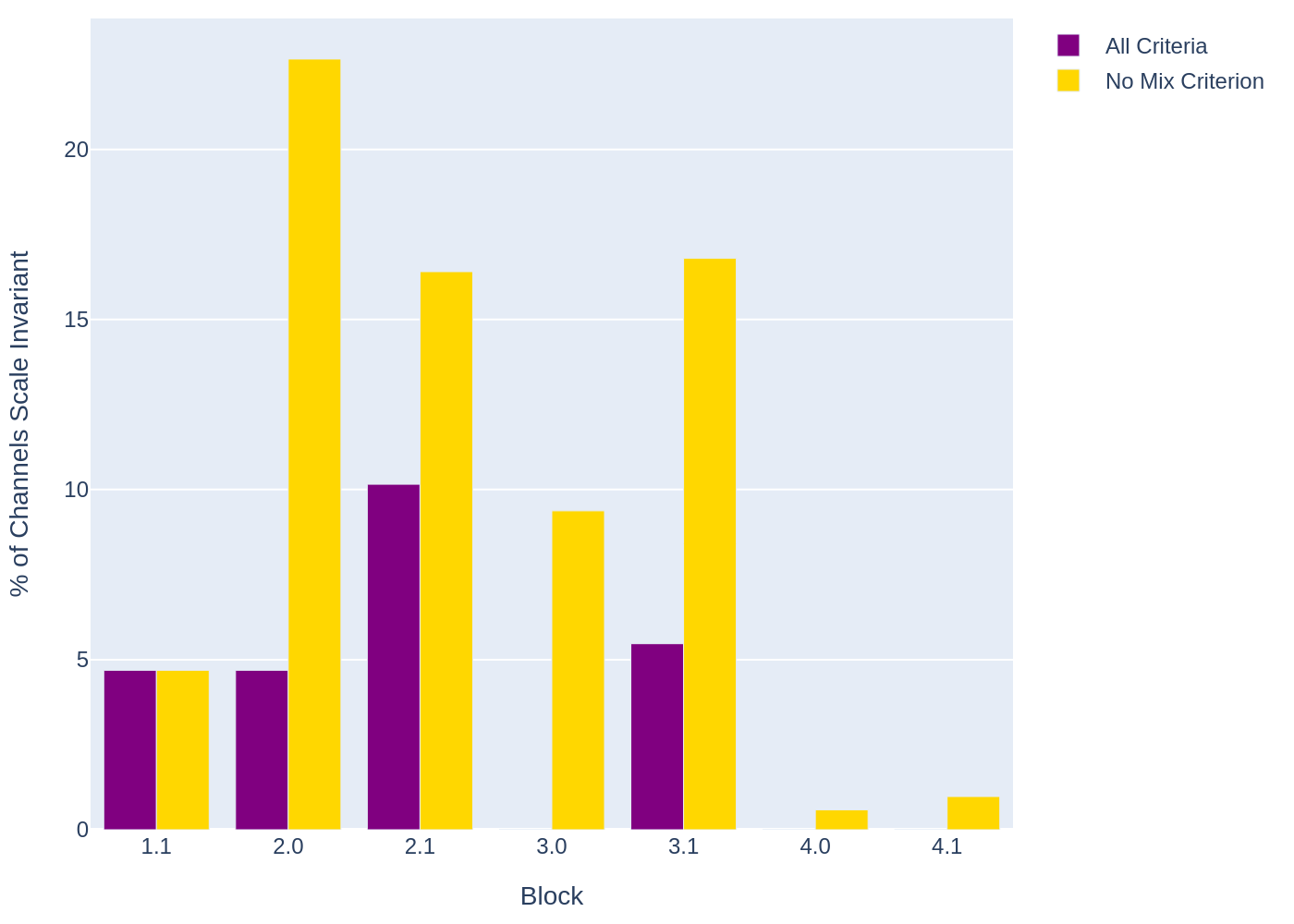}
    \caption{Percent of channels that meet all three scale invariance criteria (purple), and those with the mix ratio criterion omitted (yellow).}
    \label{fig:5}
\end{wrapfigure}  We believe more fine-grained circuit analyses of the input and block features could help clarify this issue.

\subsection{Scale Invariance in the Residual Stream}

In this section, the aforementioned three criteria (see Section \ref{scale_criteria}) are applied across all studied blocks to see if any scale invariant channels of the residual stream rise to the surface.  Indeed, in blocks 1.1, 2.0, 2.1, and 3.1, many channels fulfill the criteria.  Displayed in Figure \ref{fig:5} are the totals as a percentage of total block channels.  Zero scale invariant channels were found in blocks 3.0, 4.0, and 4.1.  At first it seemed this was due to the overwrite skew, but only a few were found in 4.0 and 4.1 even with the mix ratio criterion (Equation \ref{scale_eq:1}) removed (see yellow bars in Figure \ref{fig:5}).  This may be partly explained by the receptive fields of neurons in 4.0 and 4.1 being close to the size of the image, which constrains their features to have a fixed scale.  However, 3.0 counters this with an appreciable percentage of channels, albeit substantially less than its neighboring blocks.  It appears that overall, the reason for the dearth of scale invariance in these blocks lies more so in the absence of scaled features per se.

\begin{figure}[h]
    \centering
    \includegraphics[width=.95\linewidth]{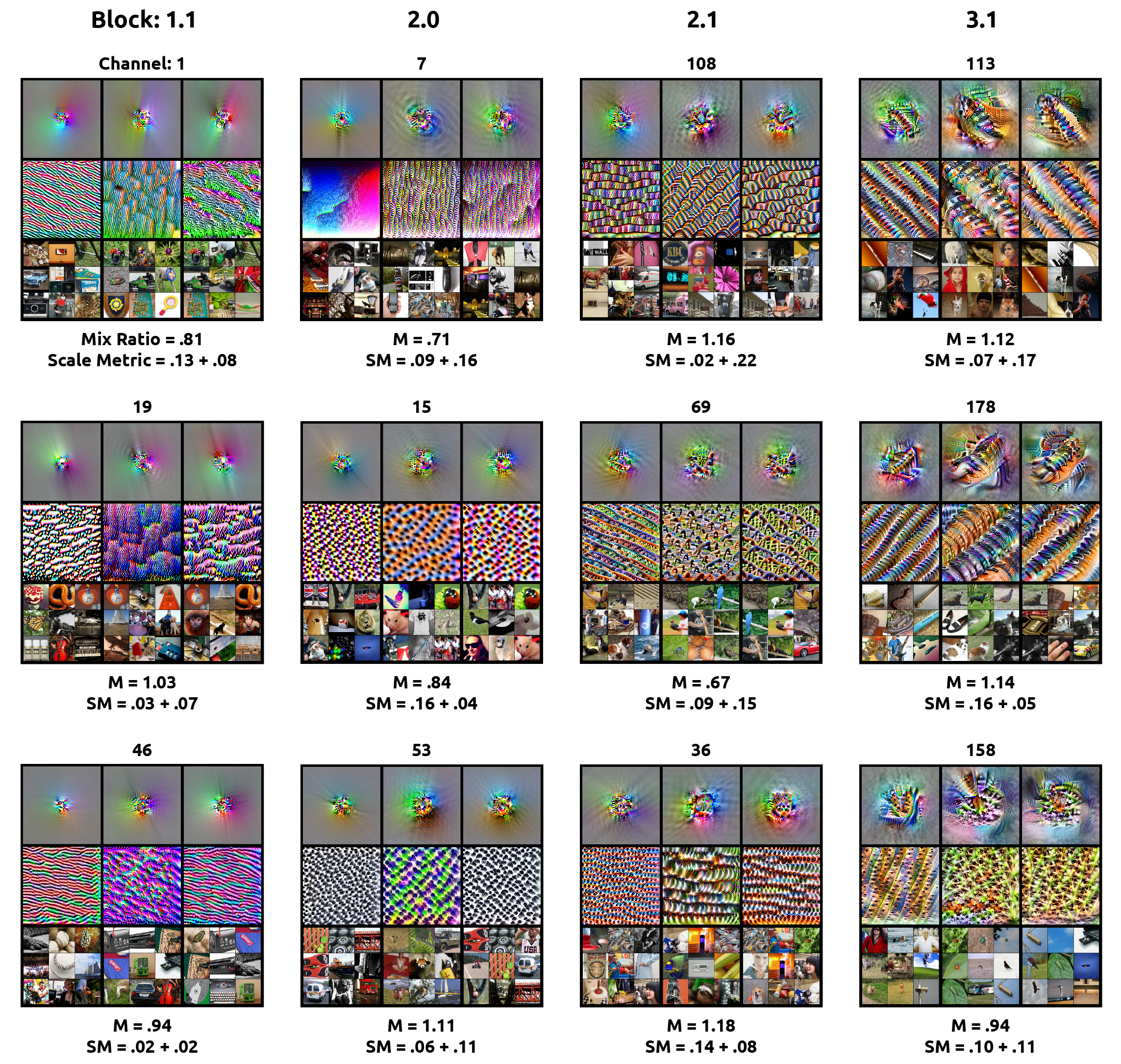}
    \caption{Columns are blocks with channels that passed the scale invariance criteria (see Section \ref{scale_criteria}).  Rows are the top 3 channels in descending order by the scale metric.  Each grid has the same layout as in Figure \ref{fig:1}.  All channel indices are zero-indexed.  Displayed for each grid are the mix ratio and the two summed deltas of the scale metric per Equation \ref{scale_eq:4}.}
    \label{fig:6}
\end{figure}

We now obtain the top 3 channels of each block with the highest scale metric \(SM\) defined in Equation \ref{scale_eq:4}.  We visualize these channels in Figure \ref{fig:6} via grids of their FZs and center-neuron top-activating ImageNet validation images, together with their mix ratio \(M\) and scale metric \(SM\).  In these examples, we uncover a variety of features such as small and large black dots (block 2.0 channel 15), rectilinear patterns (block 2.1 channel 108), and oriented cylindrical curves (block 3.1 channels 113 and 178).  Some of these examples such as the cylinders look like intuitively plausible cases of scale invariance, but we note others do not appear as obvious, such as those in block 1.1 (although there are relatively lower scale metrics for channels 19 and 46).

These results provide preliminary evidence for the existence of scale invariant features computed by the residual stream.  Our criteria may be a good starting point, but leaves more to be desired.  To add robustness to the results, we reran all three criteria but replaced FZs with the mean activation for the top 9 activating ImageNet validation images.  For all the channels which cleared both the FZ and ImageNet versions of the criteria, we visualize them in the appendix (see Section \ref{scale_inv_robust}).

\section{Discussion}

\subsection{Limitations}

As this investigation is in a nascent stage, it leaves further development to be made.  While the basic behavior of the skip-overwrite spectrum is presented, we have not addressed \textit{why} particular features are skipped, erased/overwritten, or mixed with another feature.  We also leave the overwrite skew of the downsample blocks unexplored, although the mixes visualized in appendix Section \ref{top_mixes} for 4.1 indicate significant feature convergence, making the input features redundant.  The implementation of feature erasure/overwrite also remains unexplained.  While there are hints that inhibition may be a mechanism, our results show mixed findings.  A more circuits-driven analysis may reveal if negatively-weighted features in the final layer of a block (incorporating the sign of BN's weight) bear similarities to the positively-weighted features of the input.

The criteria used to detect scale invariance might be necessary but is certainly insufficient, probably leading to false positives.  A more definitive method is needed, such as how the circuit approach reveals scale equivariance in InceptionV1 \cite{olah2020naturally}.  Similarly, if scaled versions of the same feature exist in the input and block, they must also have correspondingly scaled circuitry.  Additionally, further work could better assess if scale invariance is an explicit strategy learned in the presence of a residual stream.  An approach would be to test object recognition accuracy robustness to scale transformations against a DNN with a residual stream and an otherwise identical DNN without one.  One could also detect if scale equivariance occurs \textit{intra-layer} in the streamless DNN, which is necessary to form scale invariance in the subsequent layer in a feedforward architecture.

\subsection{Future Directions}

Aside from rectifying the limitations of this study, new directions of research can branch from this work.  We encourage the continued search for \textit{universality} in a variety of ways.  First, the skip-overwrite spectrum and scale invariance could be searched for in other residual stream architectures such as the other ResNets or vision transformers.  We consider recurrent neural networks such as CORNet-S \cite{kubilius2019brain} especially intriguing.  As weights are shared across timesteps, this allows for the detection of the identical feature across different scales, and its bypass connection can build the invariance.  Second and more speculatively, there may be commonalities in how features are updated in the residual streams of vision and language models.  For instance, a similar feature erasure/overwrite behavior in transformer-based language models has been alluded to \cite{elhage2021mathematical}.  This direction contains the possibility for the universality of residual stream management not only in different architectures, but across entirely different modalities.  Finally, as conjectured in the introduction, these findings may have relevance for visual neuroscience.  Literature has emphasized the existence of ``bypass connections'' in the visual object recognition systems of the macaque \cite{Kravitz2012-ls} and mouse \cite{Gamanut2021-az}, but little work has been done to identify the functions of these connections.  As bypass connections are analogous to the residual stream, it seems plausible that one such function is to compute scale invariant representations as described in this work.  Progress in this direction may find parallels between visual processing in artificial and biological neural networks, mechanistically explaining \textit{why} DNNs perform so well at predicting neural responses to visual stimuli in the ventral stream  \cite{Schrimpf407007}.

\subsection{Conclusion}

We have presented observations regarding the function of the residual stream in ResNet18.  We offer evidence for interpreting the residual stream as a flexible feature manager through the skip-overwrite spectrum, which also facilitates the computation of scale invariant representations.  These findings are important to further the mission of mechanistically understanding DNNs, with exciting possibilities to discover if their computations also take place in biological neural networks.

\newpage
\bibliographystyle{abbrvnat}
\bibliography{references}

\newpage
\appendix

\section{Appendix}

\subsection{Code Repository}

To promote replication and further exploration, we share our code at:

\url{https://github.com/cest-andre/residual-stream-interp}

\subsection{Feature Visualization Regularization} \label{fz_reg}

We use an identical regularization method as described in \cite{olah2017feature}, except we use different levels of jitter dependent on the depth of the layer (we still use half the initial jitter value for the second jitter transform).  This is done due to the unreliability high jitter has at producing a high activation on the center neuron once optimization has completed.  We sacrificed some interpretability that jitter offers for more accurate activation values.

We omitted jitter for the entire 1.1 and 2.0 blocks, we used jitter\(=\)4 for 2.1 second BN (\(B\)) and the entire 3.0 block, then used the default jitter\(=\)16 value for the remainder of the blocks.

\subsection{Scale Invariance Robustness} \label{scale_inv_robust}

\begin{figure}[h]
\centering
\includegraphics[width=1\linewidth]{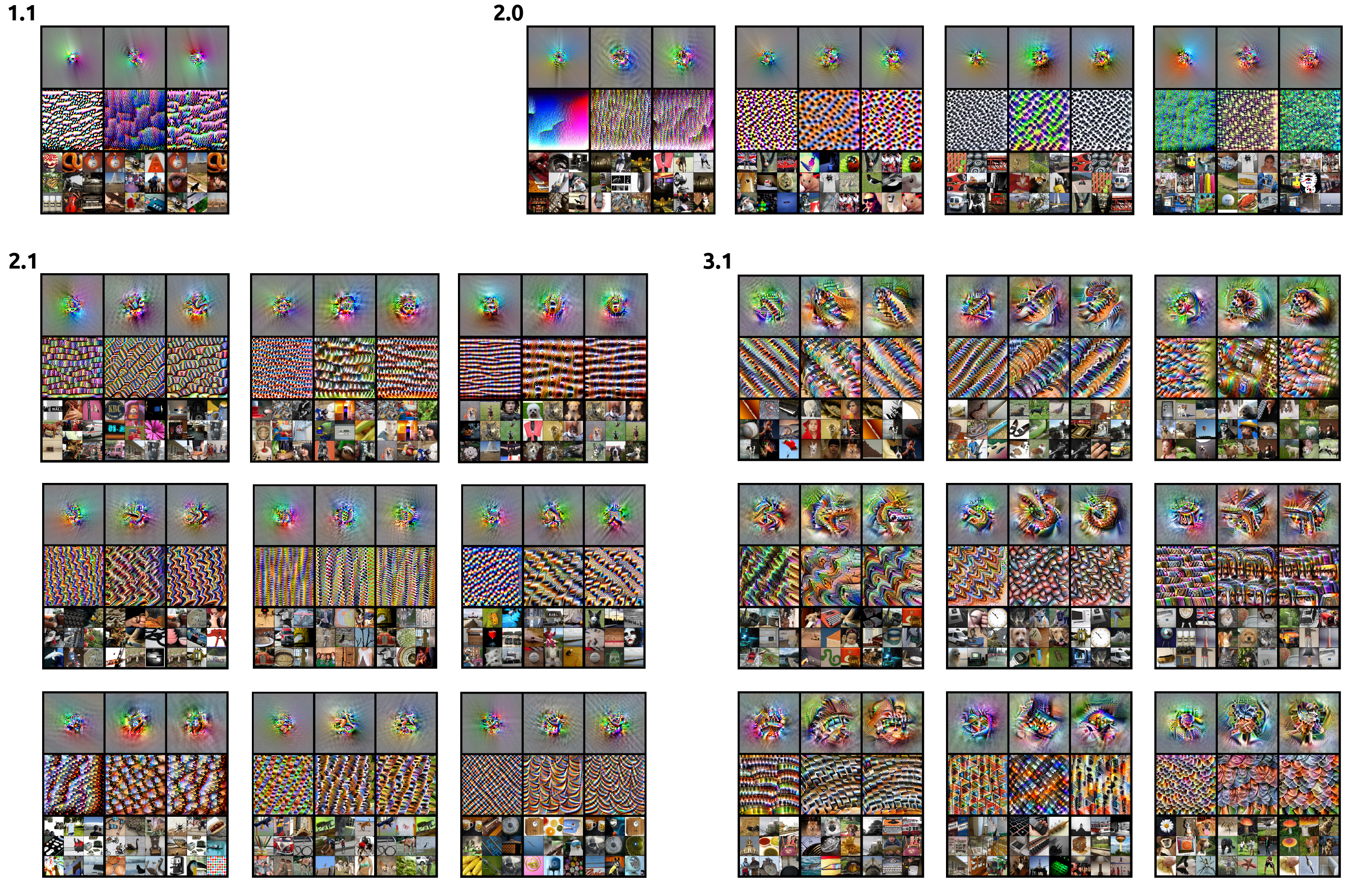}
\caption{All channels that passed the FZ and ImagetNet validation scale invariant criteria for all blocks studied (1.1 through 4.1).  The channels are ordered by the scale metric (with respect to FZ stimuli) left to right and top to bottom.  The channels are as follows (all zero-indexed):  \textbf{1.1}: 19 \quad \textbf{2.0}: 7, 15, 53, 3 \quad \textbf{2.1}: 108, 36, 13, 21, 60, 52, 93, 5, 58 \quad \textbf{3.1}: 113, 178, 22, 215, 204, 25, 29, 154, 200.}
\label{fig:7}
\end{figure}

To add robustness to our scale invariance findings, we rerun the criteria in \ref{scale_criteria}, but replace FZ stimuli \(\hat{X}_c\) with \(\textbf{V}_c\), the batch of nine ImageNet validation images that maximally activated the center neuron in the target channel \(c\).  We then replace activations \(B_c(\hat{X}_c)\) with the mean activation to this batch, \(\bar{B}_c(\textbf{V}_c)\) (same for \(I\)), for both mix ratio and scale deltas.  For all blocks studied, we display the channels that pass both this mean top validation criteria and the original FZ criteria in Figure \ref{fig:7} (we order them by the scale metric obtained from FZs).

\newpage
\subsection{Top Mixes} \label{top_mixes}

\begin{figure}[h]
\centering
\includegraphics[width=1\linewidth]{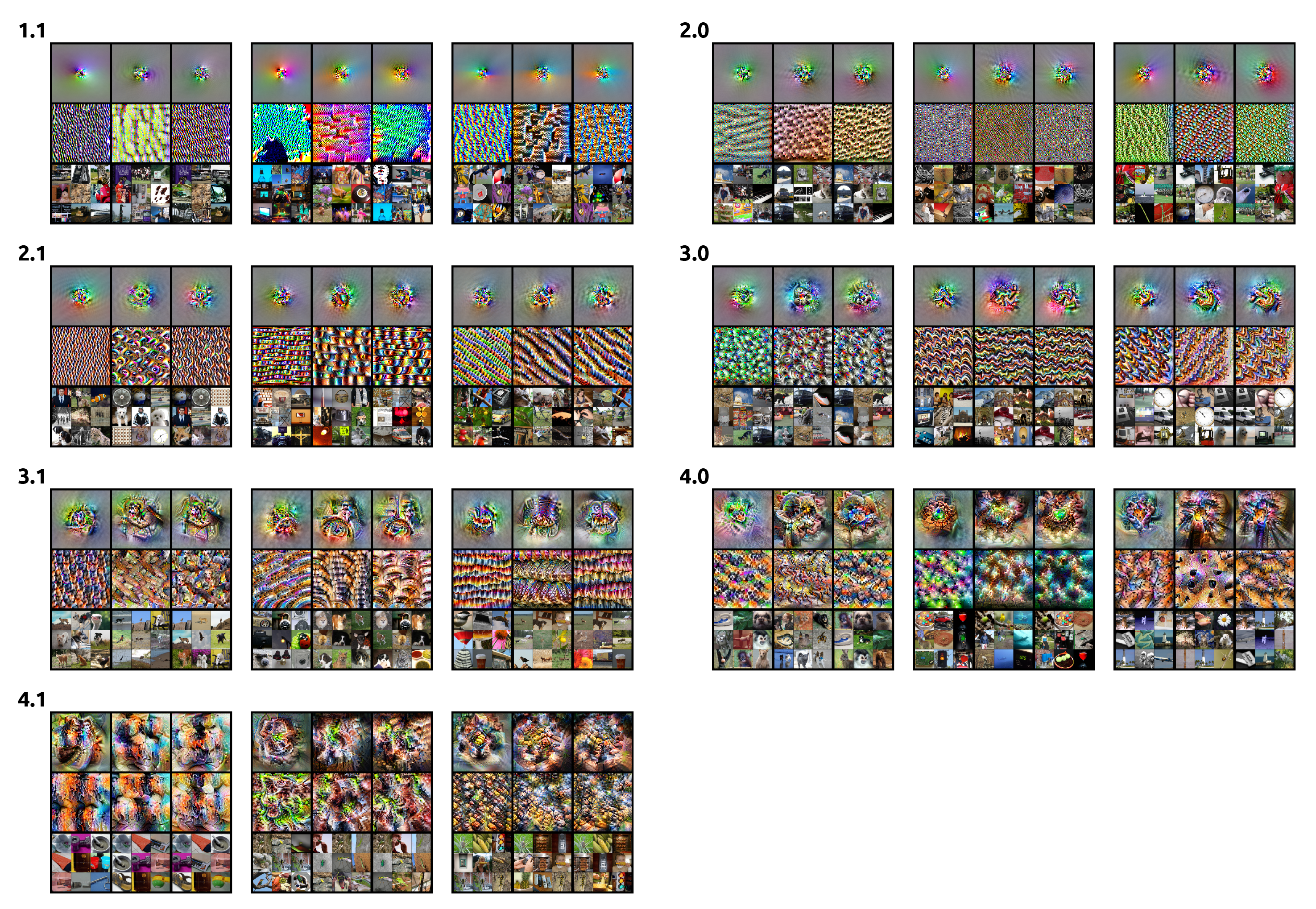}
\caption{Channels with mix ratio \(M_c\) closest to 1 in all blocks studied.  The channels are ordered left to right.  The channels are as follows (all zero-indexed): \textbf{1.1}: 44, 27, 55 \quad \textbf{2.0}: 42, 9, 78 \quad \textbf{2.1}: 22, 116, 72 \quad \textbf{3.0}: 105, 250, 204 \quad \textbf{3.1}: 128, 246, 41 \quad \textbf{4.0}: 163, 3, 51 \quad \textbf{4.1}: 154, 68, 496.}
\label{fig:8}
\end{figure}

\end{document}